%% file: ms.tex
\documentclass[lettersize,journal]{IEEEtran}
\usepackage{amsmath,amsfonts}
\usepackage{algorithmic}
\usepackage{algorithm}
\usepackage{array}
\usepackage[caption=false,font=normalsize,labelfont=sf,textfont=sf]{subfig}
\usepackage{textcomp}
\usepackage{stfloats}
\usepackage{url}
\usepackage{verbatim}
\usepackage{graphicx}
\usepackage{cite}
\usepackage{enumitem}

\usepackage{booktabs} 
\usepackage{multirow}

\usepackage{amsthm}
\usepackage{csquotes}
\usepackage{soul}
\usepackage{xcolor}
\usepackage{hyperref}

\hypersetup{
    colorlinks=true,
    urlcolor=cyan,
}

\theoremstyle{observation}
\newtheorem{observation}{Observation}[section]

\begin{document}

\title{VOICE: Variance of Induced Contrastive Explanations to quantify Uncertainty in \\ Neural Network Interpretability}

\author{Mohit Prabhushankar,~\IEEEmembership{Member,~IEEE}, Ghassan AlRegib,~\IEEEmembership{Fellow,~IEEE}

\thanks{M. Prabhushankar was with the Department
of Electrical and Computer Engineering, Georgia Institute of Technology, Atlanta,
GA, 30332 USA e-mail: mohit.p@gatech.edu.}
\thanks{G. AlRegib was with the Department
of Electrical and Computer Engineering, Georgia Institute of Technology, Atlanta,
GA, 30332 USA e-mail: alregib@gatech.edu.}
\thanks{Manuscript received - August 27, 2023; revised - May 11, 2024; accepted - May 23, 2024}}

\markboth{Journal of Selected Topics in Signal Processing}%
{Shell \MakeLowercase{\textit{Prabhushankar et al.}}: Uncertainty in Explainability}


\onecolumn 
\begin{description}[leftmargin=2cm,style=multiline]

\item[\textbf{Citation}]{M. Prabhushankar and G. AlRegib, "VOICE: Variance of Induced Contrastive Explanations to Quantify Uncertainty in Neural Network Interpretability," Journal of Selected Topics in Signal Processing (J-STSP) Special Series on AI in Signal \& Data Science, May 23, 2024.}

\item[\textbf{Review}]{Data of Initial Submission: 17 Aug 2023 \\ Date of Minor Revision: 11 May 2024 \\ Date of Acceptance: 23 May 2024}

\item[\textbf{Codes}]{\url{https://github.com/olivesgatech/VOICE-Uncertainty}}

\item[\textbf{Copyright}]{\textcopyright 2024 IEEE. Personal use of this material is permitted. Permission from IEEE must be obtained for all other uses, in any current or future media, including reprinting/republishing this material for advertising or promotional purposes,
creating new collective works, for resale or redistribution to servers or lists, or reuse of any copyrighted component
of this work in other works. }

\item[\textbf{Contact}]{\href{mailto:mohit.p@gatech.edu}{mohit.p@gatech.edu}  OR \href{mailto:alregib@gatech.edu}{alregib@gatech.edu}\\ \url{https://alregib.ece.gatech.edu/} \\ }
\end{description}

\thispagestyle{empty}
\newpage
\clearpage
\setcounter{page}{1}

\twocolumn

\maketitle

\begin{abstract}
In this paper, we visualize and quantify the predictive uncertainty of gradient-based \emph{post hoc} visual explanations for neural networks. Predictive uncertainty refers to the variability in the network predictions under perturbations to the input. Visual \emph{post hoc} explainability techniques highlight features within an image to justify a network's prediction. We theoretically show that existing evaluation strategies of visual explanatory techniques partially reduce the predictive uncertainty of neural networks. This analysis allows us to construct a plug in approach to visualize and quantify the remaining predictive uncertainty of any gradient-based explanatory technique. We show that every image, network, prediction, and explanatory technique has a unique uncertainty. The proposed uncertainty visualization and quantification yields two key observations. Firstly, oftentimes under incorrect predictions, explanatory techniques are uncertain about the same features that they are attributing the predictions to, thereby reducing the trustworthiness of the explanation. Secondly, objective metrics of an explanation's uncertainty, empirically behave similarly to epistemic uncertainty. We support these observations on two datasets, four explanatory techniques, and six neural network architectures. The code is available at \url{https://github.com/olivesgatech/VOICE-Uncertainty}

\end{abstract}

\begin{IEEEkeywords}
Predictive Uncertainty, Gradients, Contrastive explanations, Counterfactual explanations, Neural Networks, Deep Learning
\end{IEEEkeywords}

\input{Sections/1_Introduction}
\input{Sections/2_RelatedWork}
\input{Sections/3_Theory}

\input{Sections/4_Methodology}
\input{Sections/5_Results}

\input{Sections/6_Discussion}

\bibliographystyle{IEEEtran.bst}
\bibliography{references}

\begin{IEEEbiography}[{\includegraphics[width=1in,height=1.25in,clip,keepaspectratio]{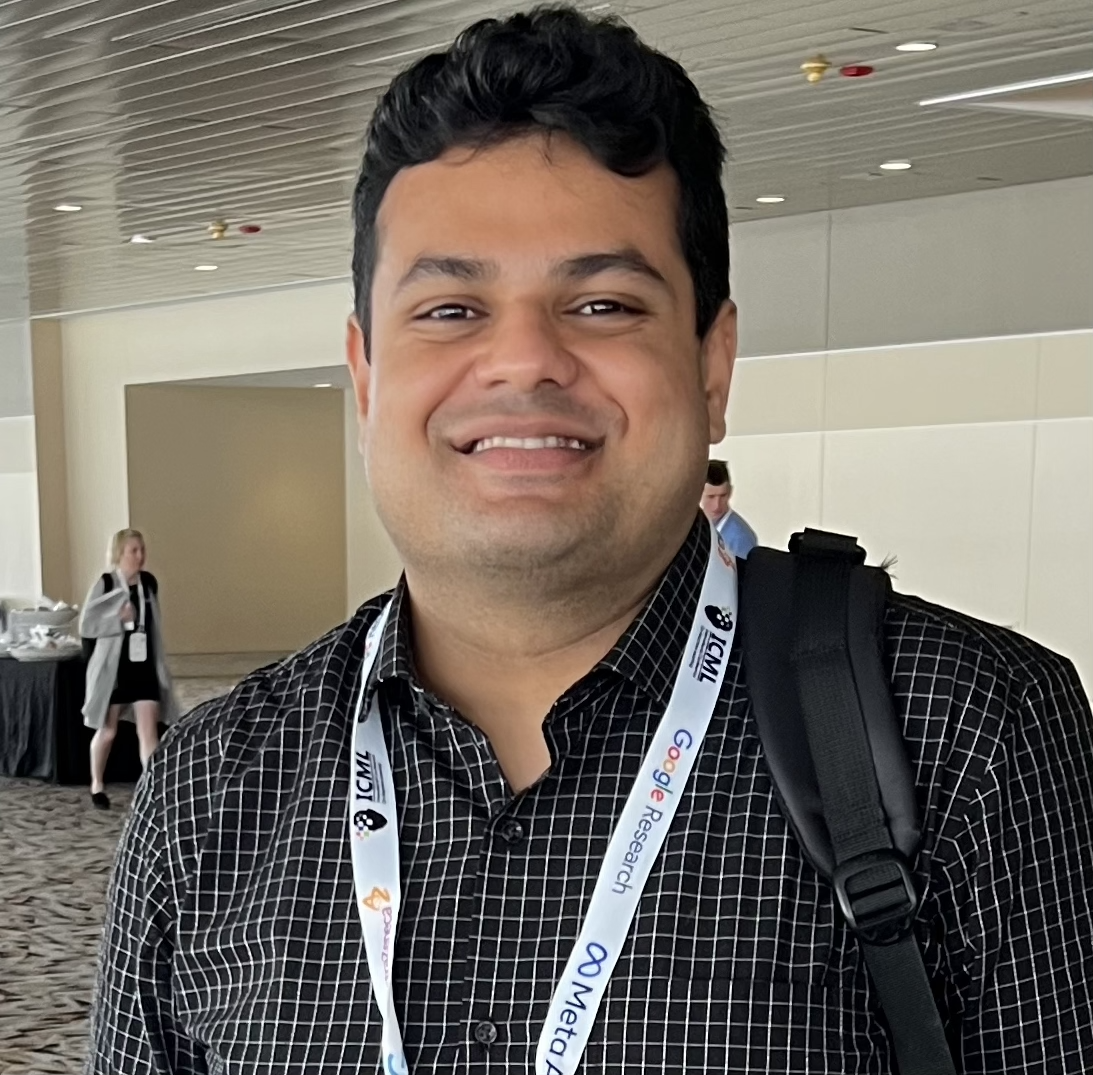}}]{Mohit Prabhushankar}
received his Ph.D. degree in electrical engineering from the Georgia Institute of Technology (Georgia Tech), Atlanta, Georgia, 30332, USA, in 2021. He is currently a Postdoctoral Research Fellow in the School of Electrical and Computer Engineering at the Georgia Institute of Technology in the Omni Lab for Intelligent Visual Engineering and Science (OLIVES). He is working in the fields of image processing, machine learning, active learning, healthcare, and robust and explainable AI. He is the recipient of the Best Paper award at ICIP 2019 and Top Viewed Special Session Paper Award at ICIP 2020. He is the recipient of the ECE Outstanding Graduate Teaching Award, the CSIP Research award, and of the Roger P Webb ECE Graduate Research Assistant Excellence award, all in 2022. He has delivered short courses and tutorials at IEEE IV'23, ICIP'23, BigData'23, WACV'24 and AAAI'24.
\end{IEEEbiography}

\begin{IEEEbiography}
[{\includegraphics[width=1in,height=1.25in,clip,keepaspectratio]{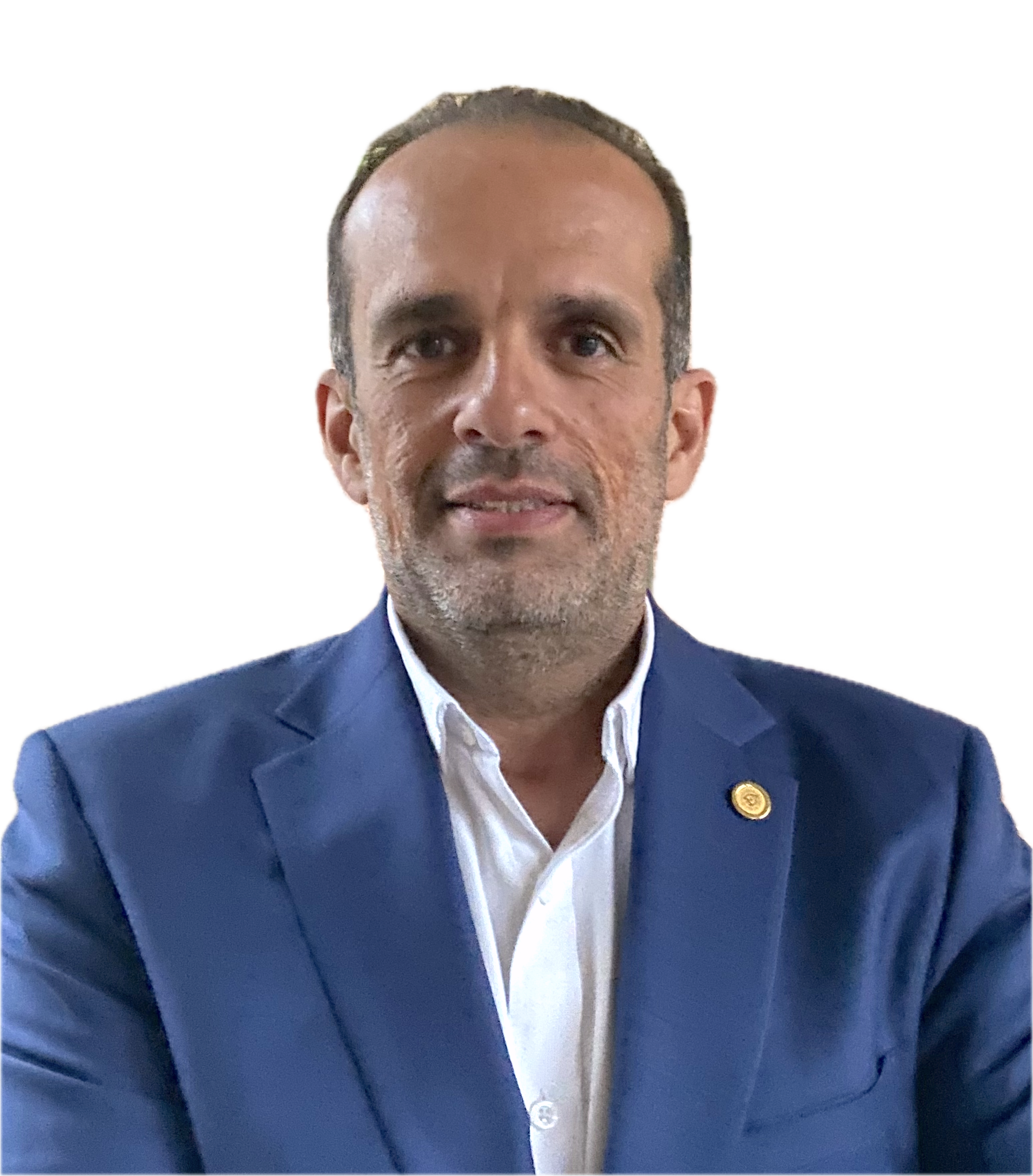}}]{Ghassan AlRegib}{\space} is currently the John and Marilu McCarty Chair Professor in the School of Electrical and Computer Engineering at the Georgia Institute of Technology. In the Omni Lab for Intelligent Visual Engineering and Science (OLIVES), he and his group work on robust and interpretable machine learning algorithms, uncertainty and trust, and human in the loop algorithms. The group has demonstrated their work on a wide range of applications such as Autonomous Systems, Medical Imaging, and Subsurface Imaging. The group is interested in advancing the fundamentals as well as the deployment of such systems in real-world scenarios. He has been issued several U.S. patents and invention disclosures. He is a Fellow of the IEEE. Prof. AlRegib is active in the IEEE. He served on the editorial board of several transactions and served as the TPC Chair for ICIP 2020, ICIP 2024, and GlobalSIP 2014.  He was area editor for the IEEE Signal Processing Magazine. In 2008, he received the ECE Outstanding Junior Faculty Member Award. In 2017, he received the 2017 Denning Faculty Award for Global Engagement. He received the 2024 ECE Distinguished Faculty Achievement Award at Georgia Tech. He and his students received the Best Paper Award in ICIP 2019 and the 2023 EURASIP Best Paper Award for Image communication Journal. 

\end{IEEEbiography}

\end{document}

%% file: Sections/1_Introduction.tex
\section{Introduction}
\label{sec:Intro}

\IEEEPARstart{V}{isual} explanations provide rationales that justify a neural network's prediction at inference by highlighting features in an image~\cite{alregib2022explanatory}. These explanations are a popular and intuitive methodology for researchers, engineers, policymakers, and users to interpret systems that use deep neural networks. The complexity of visual tasks tackled by neural networks range from microscopic textures~\cite{hu2021fabric} to earth's subsurface~\cite{shafiq2018towards}. They are used in sensitive tasks like detecting cardiovascular~\cite{poplin2018prediction} and diabetic~\cite{kokilepersaud2023clinically} risk factors in humans through retinal images. Hence, the utility of neural networks in such wide-ranging and sensitive applications call for explainability regarding their decisions. Explainability is a core tenet of neural network deployment, and is a critical field of study. A number of techniques have been proposed to explain neural network decisions at inference~\cite{selvaraju2017grad, prabhushankar2021extracting, petsiuk2018rise, springenberg2014striving, smilkov2017smoothgrad, kim2018interpretability, zhou2016learning, ribeiro2016should}. All these methods provide visual heatmaps that attribute regions or features within an image that explain the neural network's decision. The authors in~\cite{alregib2022explanatory} provide an overview of these methods by categorizing them based on design choices. One such design choice is the utility of gradients as features.

\begin{figure*}[t!]
\begin{center}
\minipage{\textwidth}%
  \includegraphics[width=\linewidth]{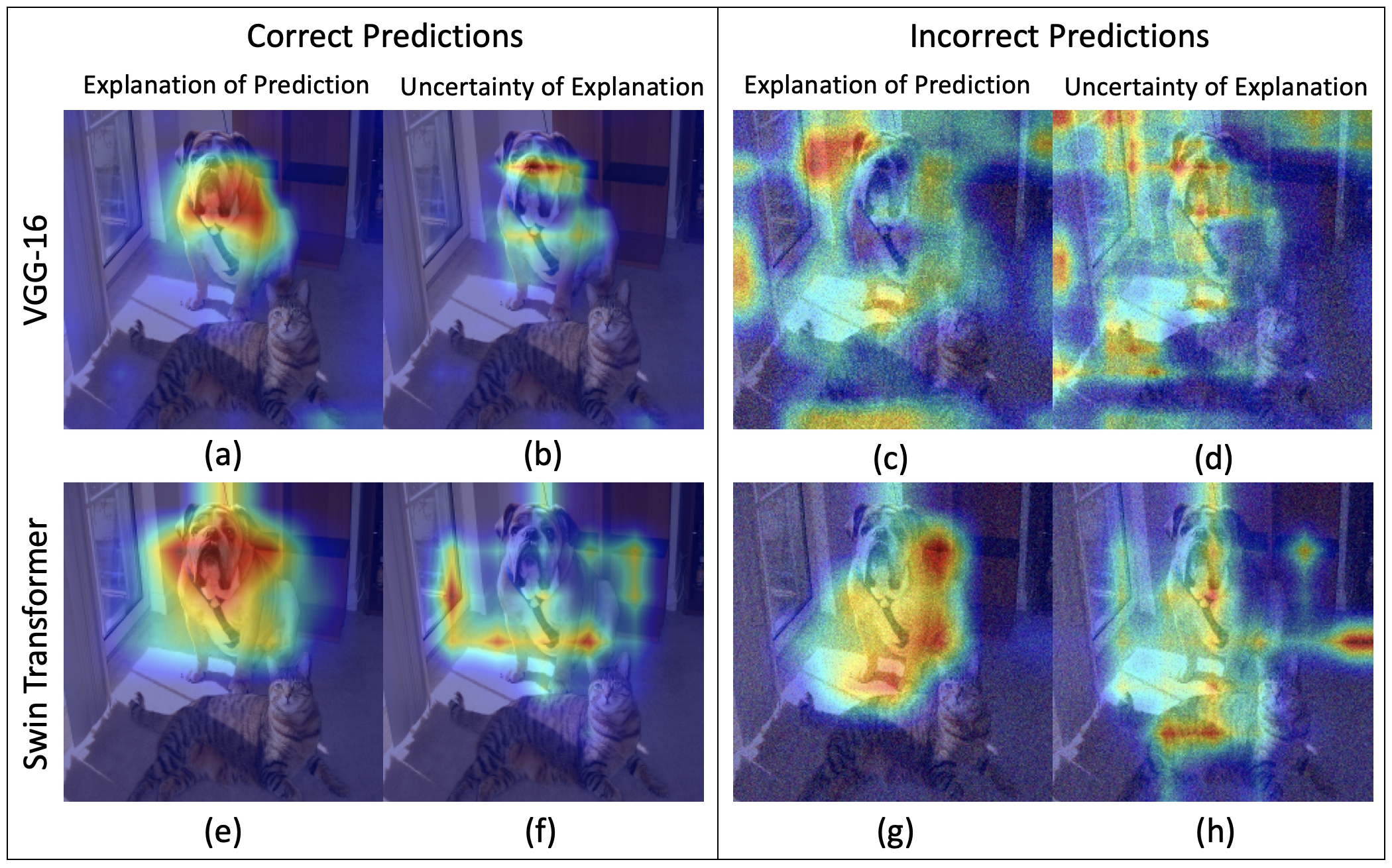}
\endminipage
\vspace{-1mm}
\caption{GradCAM~\cite{selvaraju2017grad} explanations and the proposed uncertainty visualization. Top row are results obtained on VGG-16~\cite{simonyan2014very} while bottom row are results obtained on Swin Transformer~\cite{liu2021swin}. Figs. (a), (b), (e), and (f) are obtained on a clean image where both VGG-16 and Swin Transformer predict correctly while Figs. (c), (d), (g), and (h) are results on noisy image where both networks predict incorrectly.}\label{fig:Concept}
\end{center}
\end{figure*}

Loss gradients are used in training to update network parameters towards a global minima. Gradients provide an intelligent and intuitive search direction to stochastically update the network parameters towards the minima. Recently, loss gradients have been utilized at inference to characterize data~\cite{kwon2019distorted, kwon2020backpropagated, kwon2020novelty, lee2023probing, kokilepersaud2022gradient}. Specific to explanations, a number of gradient-based methods have been proposed including GradCAM~\cite{selvaraju2017grad}, GradCAM++~\cite{chattopadhay2018grad}, Guided Backpropagation~\cite{springenberg2014striving}, and SmoothGrad~\cite{smilkov2017smoothgrad} among others. Gradient-based explanatory techniques are generally both non-explicit and non-interventionist~\cite{alregib2022explanatory}. Explicit explanatory techniques either require a change to the network architecture~\cite{zhou2016learning}, or use an approximation of the architecture~\cite{ribeiro2016should}, or require additional information to create explanations~\cite{kim2018interpretability}. Interventionist explanatory techniques require changes within data to create explanations~\cite{petsiuk2018rise}. Being non-explicit and non-interventionist allows gradient-based techniques to induce alternative paradigms of explanations. The authors in~\cite{prabhushankar2020contrastive} postulate one such alternative paradigm to propose ContrastCAM. While GradCAM backpropagates the logits of the predicted class $P$ to highlight regions that answer the question \emph{`Why P?'}, ContrastCAM backpropagates the loss between the predicted class $P$ and some contrast class $Q$ to highlight \emph{`Why P, rather than Q?'}. This paradigm of questioning is contrastive in nature. ContrastCAM induces contrastive explanations using the same methodology as the original GradCAM, but by backpropagating a loss quantity rather than the prediction. In this paper, we draw inspiration from these induced contrastive explanations to construct the uncertainty map of the base explanation. We provide a plug-in framework that works on any gradient-based neural network and explanatory technique. We further examine explainability methods and show that each explainability heatmap has an inherent uncertainty associated with it and this uncertainty is unique to the network, the image, and the technique that generated it.

In broad terms, uncertainty is the lack of knowledge within a neural network that leads to mismatched confidence at inference. For instance, neural networks are vulnerable to engineered adversarial noise~\cite{goodfellow2014explaining} where the networks predict the wrong class with high confidence. Even under pristine conditions, there is a significant gap between a network's prediction accuracy and its confidence~\cite{guo2017calibration}. Hence, a neural network does not know when to trust itself. Research in neural network uncertainty generally decomposes it into source factors~\cite{gawlikowski2023survey}. A number of techniques have been proposed to quantify uncertainties in neural networks~\cite{lakshminarayanan2017simple, gal2016uncertainty, smith2018understanding} and reduce them~\cite{prabhushankar2022introspective}. Note that all these methods quantify the uncertainty of the neural network's inference. In this paper, we visualize and quantify uncertainties of explanatory techniques that act on the model \emph{after inference}. We show one such explanatory technique, GradCAM~\cite{selvaraju2017grad}, in Fig.~\ref{fig:Concept}. In Fig.~\ref{fig:Concept}(a), GradCAM explains VGG-16's~\cite{simonyan2014very} decision of bull-mastiff by highlighting the face of the dog. The proposed uncertainty method in Fig.~\ref{fig:Concept}(b) highlights the region above and below GradCAM's explanation with an emphasis on the snout of the dog. Hence, GradCAM is uncertain of VGG-16's feature attributions based on the dog's snout and the dog's upper body. Next, GradCAM's explanation and uncertainty of Swin Transformer~\cite{liu2021swin} is visualized in Figs.~\ref{fig:Concept}(e) and (f) respectively. Notice that, similar to Fig.~\ref{fig:Concept}(b), Fig.~\ref{fig:Concept}(f) also curls around its explanation. However, unlike VGG-16, Swin Transformer is not uncertain about the dog's snout since the snout forms an important attribute within its original explanation. We further complicate this analysis by adding Additive White Gaussian Noise to the image so that both the networks predict incorrectly. The results are shown in Figs.~\ref{fig:Concept}(c), (d), (g), and (h). In contrast to the results on pristine images, we make the two following observations. Firstly, both the explanation and its associated uncertainty are more dispersed across the image. Secondly, there is a larger overlap between the explanation and its uncertainty. The second observation indicates that GradCAM is uncertain about the region it is attributing the decision to, thereby reducing the trustworthiness of the method. While the goal of the paper is to visualize the uncertainties in Figs.~\ref{fig:Concept}(b), (d), (f), and (h), we also provide two metrics based on the above observations to quantify these visualizations.

In Section~\ref{sec:Lit_Review}, we take a deeper look into existing uncertainty quantification techniques to establish the novelty of the proposed algorithm. We theoretically motivate the need for uncertainty in explainability in Section~\ref{sec:Theory}. The methodology of obtaining uncertainty visualizations as well as quantifying them is described in Section~\ref{sec:Methodology}. The results from the proposed framework are discussed in Section~\ref{sec:Results}. The key contributions of the paper include,

\begin{itemize}
    \item We propose a rigorous framework for quantifying predictive uncertainty of gradient-based visual explanations. 
    \item The proposed framework is a plug-in on top of any existing gradient-based visualization technique. 
    \item We provide two quantitative metrics to characterize the proposed uncertainty.
    \item The proposed framework allows us to conduct extensive evaluations of existing visual explanatory techniques when differentiating between correct and incorrect predictions as well as in the presence of perturbations.
\end{itemize}

%% file: Sections/2_RelatedWork.tex
\section{Related Work}
\label{sec:Lit_Review}
\subsection{Explanations and Uncertainty in Neural Networks}
\label{subsec:Reasoning}
Neural networks reason inductively, i.e. they make decisions with uncertainty which then allows speculation regarding cause~\cite{prabhushankar2021contrastive}. This is opposed to deductive reasoning where the exact nature of the cause is ascertained before making decisions. Hence, induction-based neural network reasoning allows a statistical inference that is equipped with uncertainty~\cite{lele2020should}. Explanations are a \emph{post hoc} function that attempt to justify the intractable inference of neural networks to humans. Post hoc refers to explanation functions acting on the network after the network's decision is made. Computational explanation models that attempt this justification are a function of the neural network's decision and hence the decision's uncertainty.
\subsection{Uncertainty Quantification in neural networks}
\label{subsec:Uncertainty_Review}
The science of uncertainty quantification (UQ) deals with assigning probabilities regarding decisions made under some unknown states of the system. When the system is a deep neural network, the primary research direction in UQ examines the uncertainty associated with the neural network itself. This research can be summarized by considering the sources of uncertainties, namely the data and model~\cite{gal2016uncertainty} uncertainties. A secondary research direction in UQ applies estimated uncertainty to select additional data for training~\cite{10053381}, interpreting existing results~\cite{benkert2022reliable}, estimating image quality~\cite{zhou2022ramifications}, and detecting out-of-distribution and adversarial samples~\cite{lee2022gradient}. In all cases, the statistical uncertainty of the decision is quantified. In this work, we do not estimate the network or the data uncertainty. Rather, we quantify the uncertainty of the \emph{post hoc} explanation method that acts on the network.  

Uncertainty in explainability is less researched in literature. This is partly because the goal of explainability is \emph{post hoc}, i.e. explanatory techniques act after a decision is made. On the other hand, UQ deals with estimating and reducing uncertainty of the decision itself. The authors in~\cite{patro2019u} incorporate uncertainty estimation during the training process to improve visual explanations at inference for the task of visual question answering. In~\cite{chowdhury2023explaining}, causal metrics of necessity and sufficiency are used to evaluate the uncertainty of explanations in tabular data. The authors in~\cite{slack2021reliable} propose an uncertainty framework using explanations. They propose BayesLIME and BayesSHARP by measuring the changes in the disagreement between explanations with slight perturbation. The idea of disagreement as a measure of uncertainty has been explored in multiple domains~\cite{bomberger1996disagreement, zhou2022ramifications}. In this paper we show that each individual perturbation itself creates uncertainty within an explanatory framework and that the explanatory technique has its own uncertainty. To do so, we consider a combination of the data and model uncertainties, which is termed as the predictive uncertainty~\cite{kendall2017uncertainties}.

Predictive uncertainty is measured as the \textit{variance} of the model outputs~\cite{mckay1995evaluating}. The authors in~\cite{kohler2019uncertainty} discuss three  methods to estimate predictive uncertainty estimates: Deep Ensembles \cite{lakshminarayanan2017simple}, MonteCarlo dropout (MC-dropout \cite{gal2016dropout}), and a combination of both \cite{smith2018understanding}. The variance for a prediction is calculated across multiple model outputs to obtain uncertainty. Predictive uncertainty techniques are generally evaluated against log-likelihood and brier score. In this paper, we utilize the definition of predictive uncertainty to estimate the uncertainty of explanation techniques.
\subsection{Visual Explanations}
\label{subsec:Visual_Explanations}
A popular means of visual explanations are attribution masks as shown in Fig.~\ref{fig:Concept}. These masks highlight features that led a neural network to make its decision. The authors in~\cite{alregib2022explanatory} provide a taxonomy of explanatory techniques based on the design choices of the explanatory techniques and expand on the advantages of gradient based explanatory methods. Popular gradient-based methods including GradCAM~\cite{selvaraju2017grad}, GradCAM++~\cite{chattopadhay2018grad}, Guided Backpropagation~\cite{springenberg2014striving}, and SmoothGrad~\cite{smilkov2017smoothgrad} all function by backpropagating the prediction logit and either directly visualizing some processed gradients~\cite{springenberg2014striving, smilkov2017smoothgrad} or using gradients as a weighing function on other features~\cite{selvaraju2017grad, chattopadhay2018grad}. Furthermore, the authors in~\cite{prabhushankar2020contrastive} posit a psychological definition to explanations whereby all the above feature attribution methods answer \emph{`Why P?'} questions where $P$ is the prediction. This leads to other paradigms of explanations namely contrastive and counterfactual explanations. Specifically, contrastive methods answer the question \emph{`Why P, rather than Q?'}. In this explanatory paradigm, the network highlights features that separate the prediction $P$ from a contrast class $Q$.~\cite{prabhushankar2020contrastive} propose ContrastCAM by backpropagating a loss function between $P$ and $Q$, instead of the prediction logit through any gradient-based \emph{`Why P?'} framework. However, a number of recent works have questioned the validity of these gradient-based visual explanations. The authors in~\cite{nie2018theoretical} suggest that all gradient-based visualization techniques partially recover the input image. In this paper, we show that it is the evaluation of any gradient-based visual explanation that lends itself to reducing the predictive uncertainty in a neural network. In other words, explanations highlight features that reduce the predictive uncertainty of the neural network. We describe the evaluation of explanations in literature before discussing its implications on predictive uncertainty in Section~\ref{sec:Theory}.\vspace{-2mm}
\subsection{Evaluation of visual explanations}
\label{subsec:Eval_Explanations}
A number of techniques have been proposed for evaluating visual explanations. Specifically, the authors in~\cite{alregib2022explanatory} provide a taxonomy of explanatory evaluation. The first evaluation strategy, utilized in multiple explanation methods, is direct human evaluation. Qualitative results from explanatory techniques $\mathcal{M}_1$ and $\mathcal{M}_2$ are shown to humans in controlled scenarios and their preferences between the methods are noted. However, large scale experiments involving humans is expensive and time consuming. Hence, an indirect and targeted class of strategies is utilized to evaluate explanations.

Indirect and targeted evaluation techniques involve masking a given image using an explanatory map and checking for accuracy of the masked image on the model. If the decision is unchanged, then the explanation has captured the essential features, hence fulfilling the requirements of being a \emph{post hoc} explanation. A number of works propose different methodologies of masking. For instance, the authors in~\cite{petsiuk2018rise} mask pixel by pixel, by using the explanatory heatmap as a probability map. They mask the pixels in the original image based on descending order of probability. The authors in~\cite{prabhushankarCausal} disagree with the assumption of pixelwise masking and instead mask structure-wise, since the human vision system is attuned to structures rather than pixels. They use Huffman encoding as a proxy for structures and mask the unimportant structures based on explanations. The authors in~\cite{chattopadhay2018grad} directly threshold the mask and only pass those regions in the input image which are above the threshold value. In Section~\ref{sec:Theory}, we show that this masking creates a subset of selected features, which when passed through the network, creates its own unique uncertainty.\vspace{-2mm}
\subsection{Gradients as uncertainty features}
\label{subsec:Gradients}
Gradients provide a measurable change to the network parameters. During training, a loss function is backpropagated across the network and the network parameters are updated until a local minima is reached~\cite{rumelhart1986learning}. Recently, a number of works have used gradients as features to characterize data as a function of network weights. This characterizations has shown promising results in a number of disparate applications including novelty~\cite{kwon2020novelty}, anomaly~\cite{kwon2020backpropagated}, and adversarial image detection~\cite{lee2022gradient}, severity detection~\cite{kokilepersaud2022gradient}, image quality assessment~\cite{kwon2019distorted}, and human visual saliency detection~\cite{sun2020implicit}. A number of theories as to their efficacy has been put forward including neurobiological~\cite{prabhushankar2023stochastic}, behavioral~\cite{prabhushankar2022introspective}, and reasoning-based~\cite{prabhushankar2021contrastive}. In this paper, we adopt the interpretation of gradients as encoding the uncertainty of a loss function~\cite{lee2023probing, lee2020gradients}. This uncertainty is reflected in all explanatory techniques that backpropagate loss, i.e. gradient-based explanatory methods.

%% file: Sections/3_Theory.tex
\section{Theory}
\label{sec:Theory}
\subsection{Visual Explanations}
\label{subsec:Explanations_theory}
In mathematical notations, given a trained neural network $f: \mathcal{X} \rightarrow \mathcal{Y}$ and an input $x \in \mathcal{X}$, a \emph{post hoc} explanation $\mathcal{M}(\cdot)$ highlights features $\mathcal{T}_P$ that lead to a decision $P\in \mathcal{Y}$. This is a probabilistic definition of \emph{observed correlation} explanation as provided in~\cite{alregib2022explanatory}. Observed correlation explanations answer the question \emph{`Why P?'}. The explanatory technique $\mathcal{M}(\cdot)$ is a function of the prediction $P$, the input $x$, the network $f(\cdot)$, and the methodology used to derive the explanation, denoted by $m$. Hence, an explanation is represented as $\mathcal{M}_m(f, x, P)$. Fig.~\ref{fig:Concept}(a) represents $\mathcal{M}_{GradCAM}(\text{VGG, cat-dog, bull mastiff})$, Fig.~\ref{fig:Concept}(e) is $\mathcal{M}_{GradCAM}(\text{Swin Transformer, cat-dog, bull mastiff})$, Fig.~\ref{fig:Concept}(c) depicts $\mathcal{M}_{GradCAM}(\text{VGG, cat-dog, coral reef})$ and Fig.~\ref{fig:Concept}(g) is $\mathcal{M}_{GradCAM}(\text{Swin, cat-dog, Border terrier})$. Note that the difference in $y$ in Fig.~\ref{fig:Concept}(c) and (g) is because VGG and Swin predicted $x$ as coral reef and border terrier respectively. Similarly, Fig.~\ref{fig:Explanations_Correct} shows the results when varying $m$ between GradCAM, GradCAM++, Guided Backpropagation, and SmoothGrad and $x$ between random images from ImageNet. 

Alternative to \emph{observed correlation} explanations are \emph{observed contrastive} explanations~\cite{alregib2022explanatory}. Contrastive explanations are `\emph{Why P, rather than Q?}' explanatory techniques that provide contextual explanations that highlight features for why the network predicted class $P$ rather than some other class $Q$. The authors in~\cite{prabhushankar2020contrastive} provide a plug-in approach that induces gradient-based correlation explanations to produce contrastive explanations. They backpropagate a loss $J(P,Q)$ instead of the logit $y_P$ within $\mathcal{M}_m(f, x, P)$. Hence, these explanations can be represented as $\mathcal{M}_m(f, x, J_{P,Q})$. In this paper, we make use of these plug-in methods to derive uncertainty of the base correlation explanatory maps $\mathcal{M}_m(f, x, P)$.

\subsection{Predictive Uncertainty Decomposition}
\label{subsec:Var_Decomposition}
Under the assumption of a fixed model, predictive uncertainty is defined as the variability in the model prediction associated with changing the input values~\cite{mckay1995evaluating}. In a neural network, predictive uncertainty is the variance in the probability distribution of the outputs of the network given by $f(x)$. Hence, the predictive uncertainty is given by $V[f(x)]$ where $V[\cdot]$ is the variance of the predicted logit probabilities. For ease of notation, we refer to $f(x)$ as the output $y$. The author in~\cite{mckay1995evaluating} decomposes the variance of logit probabilities as,
\begin{equation}\label{eq:Decomposition}
    V[y] = V[E(y|S_x)] + E[V(y|S_x)].
\end{equation}
where $S_x$ is some perturbation of the input $x$. In this paper, $S_x$ is the explanation-masked input image given by $S_x = \mathcal{M}_m(f,x,P) \odot x$. $\odot$ is the element-wise Hadamard product. The first term of the RHS in Eq.~\ref{eq:Decomposition} is the variance in the expectation of $y$ under explanation masked image $S_x$. Note that current explanatory techniques use $V[E(y|S_x)]$ as a means of evaluating their methods. The explanations, $\mathcal{M}_1$ and $\mathcal{M}_2$, mask (or perturb) the image and are passed back in the network to show that the prediction $y$ does not change in one of the methods. Mathematically, this equals $E(y|S_{x1}) > E(y|S_{x2})$, where $S_{x1} = \mathcal{M}_1(f,x,y) \odot x$ and $S_{x2} = \mathcal{M}_2(f,x,y) \odot x$. Hence this leads to our first observation.
\begin{observation}\label{obs:partial}
Visual explanations are evaluated to (partially) reduce the predictive uncertainty in a neural network.
\end{observation}
We say partially since from Eq.~\ref{eq:Decomposition}, only the first term $E(y|S_x)$ is used for evaluation and its variance is reduced. The second term is the residual between the chosen $S_x$ and all other possible $S_{x}'$ where $x = S_x \cup S_{x}'$. In other words, the features \emph{not} chosen by the explanation $\mathcal{M}(\cdot)$ that are present in $S_{x}'$ are responsible for the predictive uncertainty $V[y]$. A demonstration of this is in the results of uncertainty in Fig.~\ref{fig:Concept}(b). The network is uncertain regarding its classification based on the snout of the dog since it shares this feature with other dog breeds in the dataset. However, GradCAM $m(\cdot)$ does not consider the snout as salient as the jowls of the dog for its decision. In our proposed method, we consider the second residual term given by $E[V(y|S_x)]$ to calculate this uncertainty. This leads to our second observation.
\begin{observation}\label{obs:function}
Uncertainty in explainability occurs due to all combinations of features that the explanation did not attribute to the network's decision.
\end{observation}
The simple methodology for evaluating $V(y|S_x)$ is by changing $S_x$ randomly to obtain $S_x'$ in a Monte-Carlo fashion. However, all combination of features on complex data like images are long and impractical~\cite{lopez2017discovering}. We propose a methodology that effectively computes this term.
\subsection{VOICE Theory}
\label{subsec:VOICE_Theory}
The second variance term in Eq.~\ref{eq:Decomposition} $V(y|S_x)]$ is the variability in $y$ under the presence of residual $S_x'$. For a large majority of changes in $S_x$, there is no change in $y$, leading to $V(y|S_x') = 0$. Hence, Monte-Carlo sampling of $S_x'$ can be ineffective for a well trained network. Instead, we use contrastive explanations that target changes within $x$ that lead to change in prediction from $P$ to $Q$. In other words, contrastive explanations highlight those feature $S_{P,Q}$ that differentiate prediction $P$ from $Q$. The prediction $y$ changes when the features $S_{P,Q}$ are deleted from $S_x$. Mathematically the changed probability $y'$ is given by,
\begin{align}
    y' &= f(S_x')\\
    S_x' &= S_x - S_{P,Q}
\end{align}
Therefore, we induce changes in $y$ by obtaining contrastive explanations and taking a variance across the features $S_x'$ that lead to a change in outputs $y$. We term this methodology as VOICE: Variance Of Induced Contrastive Explanations. The exact procedure for generation of VOICE uncertainty is described in Section~\ref{sec:Methodology}.

%% file: Sections/4_Methodology.tex
\begin{figure*}[t!]
\begin{center}
\minipage{\textwidth}%
  \includegraphics[width=\linewidth]{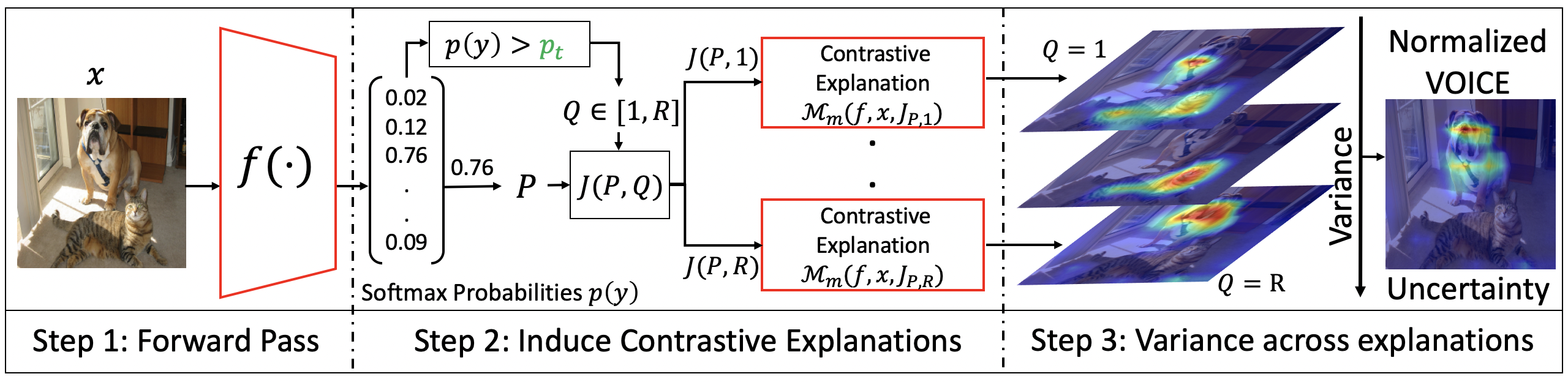}
\endminipage
\vspace{-1mm}
\caption{Proposed VOICE uncertainty generation explained using three steps. A prediction $P$ is made in Step 1. Contrastive explanations are induced using $R$ contrast classes in Step 2. $R$ is the number of class probabilities that  exceed a threshold $p_t$. In Step 3, variance across vertically stacked explanations is calculated pixelwise and normalized to produce VOICE uncertainty. The red boxes are plug-in frameworks. Green $p_t$ is a hyperparameter.}\label{fig:Block_Diagram}
\end{center}
\end{figure*}
\section{Methodology}
\label{sec:Methodology}
In this section, we describe the methodology for obtaining VOICE along with some objective metrics to characterize VOICE uncertainty.

\subsection{VOICE: Variance Of Induced Contrastive Explanations}
\label{subsec:VOICE}
The block diagram for obtaining VOICE uncertainty is shown in Fig.~\ref{fig:Block_Diagram}. Step 1 in Fig.~\ref{fig:Block_Diagram} involves a forward pass of an image $x$ to obtain a prediction $P$ from a trained network $f(\cdot)$. Existing explanatory techniques $m_m(f, x, P)$ where $m$ is any gradient-based method acts on $P$ to answer the question `\emph{Why P?}' by highlighting visual features that led to the decision $P$. Note that any explanatory method that backpropagates the logit $y_P$ of the prediction $P$ is a necessary requirement for our method. This is because in Step 2, a contrastive explanation is induced by backpropagating a loss function $J(\cdot)$ between a prediction $P$ and some contrast class $Q$. In this paper, we use cross entropy loss since it is commonly used in training the network. Note that there are $N$ possible contrast classes to create contrastive explanations where $N$ is the number of classes the network is trained to discriminate against. In ImageNet, $N$ is $1000$. Qualitatively, each contrastive explanation answers the question \emph{'Why P, rather than Q?'}. For a well trained network, the softmax probabilities of only a subset of the $1000$ classes are above a threshold $p_t$. Let $R$ denote these classes. In step 2, we backpropagate a loss $J(P,Q)$ through the explanatory methods $m_m(f, x, J_{P,Q})$ where the loss replaces the prediction logit $y_P$ similar to Contrast-CAM from~\cite{prabhushankar2020contrastive}. We do this for $R$ classes to obtain $R$ contrastive explanations. In Step 3, the $R$ induced contrastive explanations are vertically stacked, and their variance is taken across each pixel to obtain a single heatmap. This heatmap is normalized to obtain the Variance Of Induced Contrastive Explanations (VOICE) uncertainty map $u_m$, that is a function of the network $f(\cdot)$, the image $x$, the decision $P$ and the explanatory technique $m(\cdot)$.

\begin{figure*}[t!]
\begin{center}
\minipage{\textwidth}%
  \includegraphics[width=\linewidth]{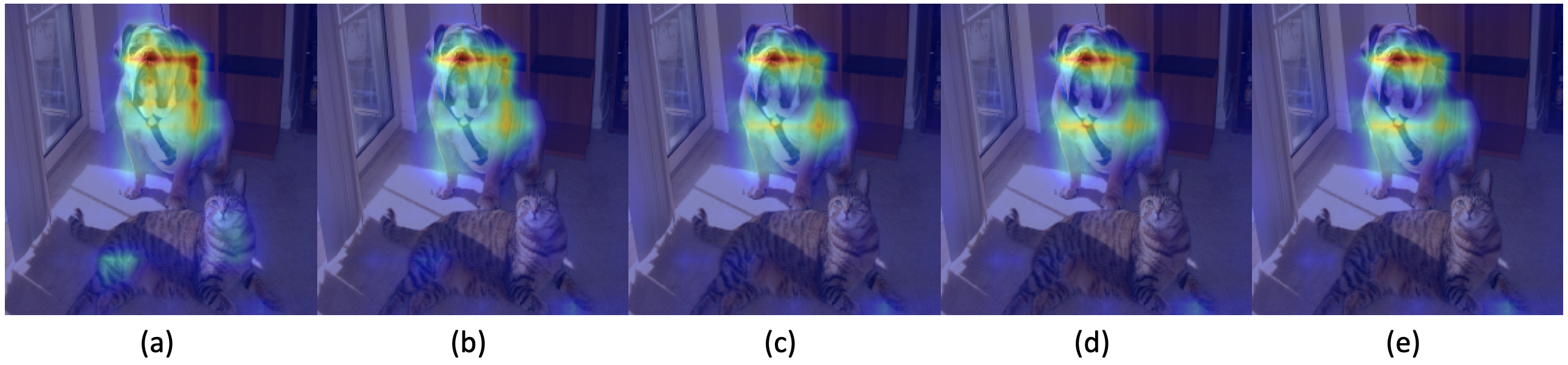}
\endminipage
\vspace{-1mm}
\caption{Proposed VOICE uncertainty on GradCAM~\cite{selvaraju2017grad} explanation with multiple threshold probabilities $p_t$ with (a) $p_t = 0.001$, (b) $p_t = 0.0001$, (c) $p_t = 0.00001$, (d) $p_t = 0.000001$, (e) $p_t = 0.0000001$.}\label{fig:Threshold}
\end{center}
\end{figure*}

\noindent\textbf{Choice of $R$: }In Fig.~\ref{fig:Threshold}, we show VOICE uncertainties on the cat-dog image from ImageNet when it is passed through VGG-16 and explained using GradCAM at different probability thresholds $p_t$. Fig.~\ref{fig:Threshold}(a) with $p_t = 0.001$ represents the case when only those contrastive maps whose class probabilities have a greater than random chance $\big(\frac{1}{N}\big)$ are induced and used, where $N = 1000$ is the number of classes in ImageNet dataset. Progressively increasing the threshold leads to a better uncertainty heatmap until $p_t = 0.00001$ in Fig.~\ref{fig:Threshold}(c), after which there is no visual difference. In this paper, we use $p = 0.00001$ for all visualizations.

\subsection{Quantitative Measures on VOICE}
\label{subsec:Quant_Measures}
Variance across the contrastive explanations provides a heatmap that can be visualized. This is shown in Fig.~\ref{fig:Concept}. In this subsection, we provide objective measures to quantify the uncertainty map. We provide two measures, one of which is a full reference metric, i.e. needing access to the base explanatory map $m$ to quantify $u_m$, and another is a no-reference metric, i.e. does not require access to $m$ to quantify $u_m$. The terminologies of full-reference and no-reference are derived from the image quality assessment literature~\cite{temel2016unique}.

\paragraph{Intersection over Union (IoU)} The explanatory map $m$ are the feature attributions that lead to a decision $P$. If the explanation-specific uncertainty map $u_{m}$ overlaps with the explanation $m$, then the overlap can be interpreted as the network being uncertain about the regions in the image that it is using to make its decision. Hence, with a high overlap, networks do not trust their own decisions. We quantify this overlap using Intersection over Union (IoU) metric, calculated as,
\begin{equation}\label{eq:IoU}
    IoU_{t} = \frac{|u_{m} \cap m|}{|u_{m} \cup m|},
\end{equation}
where the LHS is the intersection over union given a threshold $t$. The numerator of the RHS denotes the number of pixels where $u_{m}$ and $m$ overlap and the denominator signifies the number of pixels which forms the union of $u_{m}$ and $m$. Higher $IoU_t$, less trust can be placed on the attributions made by the explanatory map. We show this in Section~\ref{sec:Results} where wrong predictions made by networks generally have higher IoU compared to correct predictions. Additionally, since both $m$ and $u_{m}$ are required in Eq.~\ref{eq:IoU}, it is a full-reference metric.

\paragraph{Signal to Noise Ratio (SNR)} Signal-to-Noise Ratio is a metric in image processing that is used to compare the intensity of a desired signal to the intensity of background noise. SNR is generally calculated as,
\begin{equation}\label{eq:SNR}
    SNR = \frac{\mu(u_m)}{\sigma(u_m)},
\end{equation}
where $u_m$ is the uncertainty map, $\mu$ is the expected value or mean of the uncertainty map and $\sigma$ is the standard deviation of the uncertainty map. SNR expressed as Eq.~\ref{eq:SNR} is the inverse of the coefficient of variation (CV). CV is a standardised measure of the dispersion of a probability distribution. Hence, in the context of uncertainty quantification, SNR measures the ratio of the mean intensity of $u_m$ against the dispersion of the intensity of $u_m$ across the pixels. Higher the SNR of $u_m$, more is the uncertainty. Therefore, for a good explanatory map, SNR of its uncertainty must be low. Note that unlike IoU, SNR is calculated using $u_m$ only without needing access to $m$ and hence, is a no-reference metric.

In literature, log-likelihood and brier score are used as a proxy to quantify uncertainty~\cite{gal2016uncertainty}. The authors in~\cite{zhou2022ramifications} note that both brier score and log-likelihood metrics require ground truth annotations for their calculation. However, under label uncertainty in practical scenarios of biomedical diagnosis, these annotations are sometimes unavailable. Hence, the proposed methodology provides a means to approximate uncertainty without requiring ground truth. In Section~\ref{sec:Results}, we show that both SNR and IoU are correlated with log-likelihood measure since they measure the residual term of predictive uncertainty from Eq.~\ref{eq:Decomposition}.

%% file: Sections/5_Results.tex
\begin{figure*}
\begin{center}
\minipage{\textwidth}%
  \includegraphics[width=0.99\linewidth]{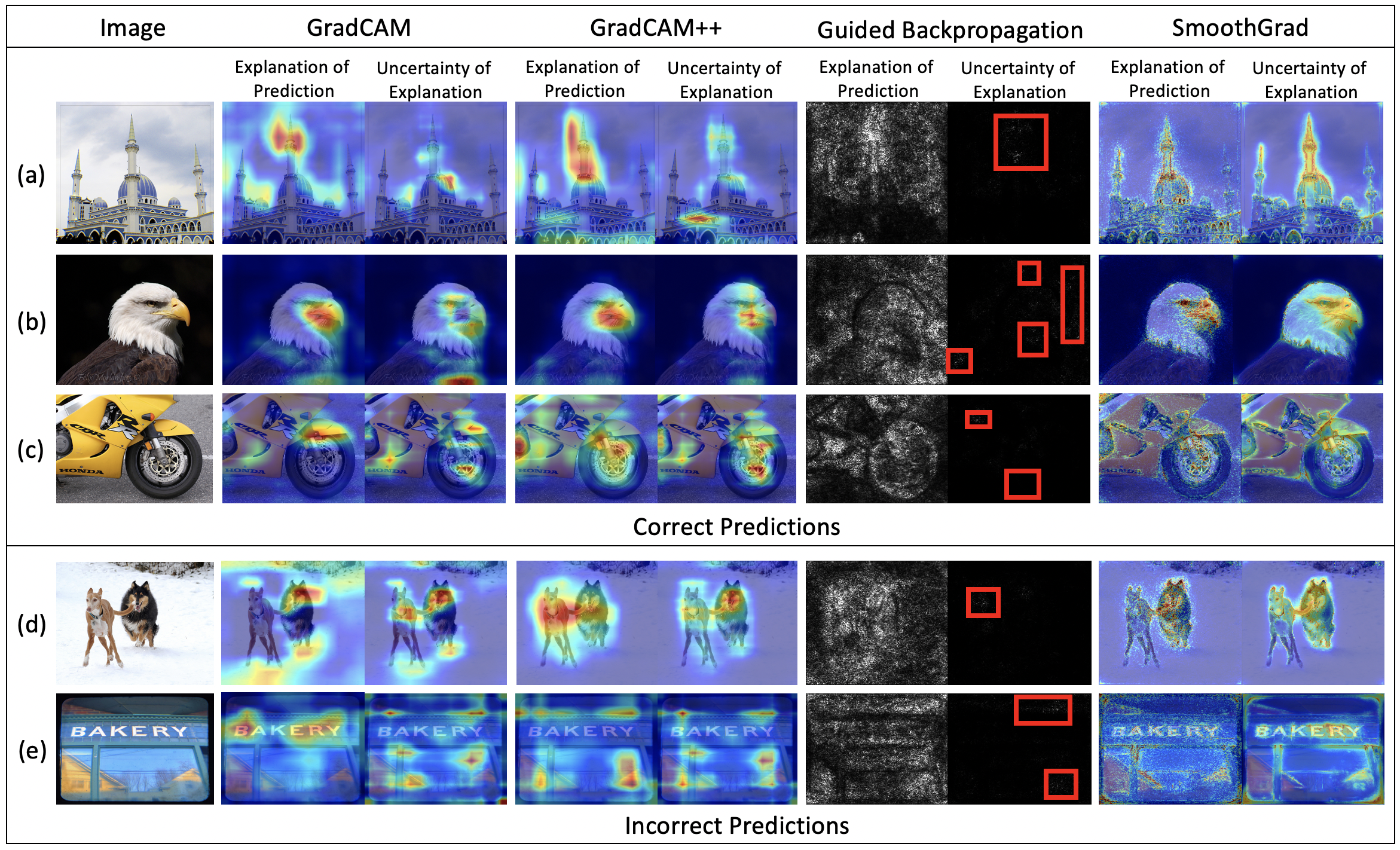}
\endminipage
\vspace{-1mm}
\caption{Visualization of GradCAM, GradCAM++, Guided Backpropagation, and SmoothGrad explanations and their corresponding uncertainties on three correctly classified and two incorrectly classified randomly selected images from ImageNet validation set. Guided Backpropagation uncertainty is not easily visible and red boxes are used to show highlights.}\label{fig:Explanations_Correct}
\end{center}
\end{figure*}

\section{Results}
\label{sec:Results}

\input{Table/tab_Results}
In this section, we show both qualitative and quantitative results for VOICE uncertainty for the application of image recognition. We use four popular gradient based explanatory techniques including GradCAM~\cite{selvaraju2017grad}, GradCAM++~\cite{chattopadhay2018grad}, Guided Backpropagation~\cite{springenberg2014striving}, and SmoothGrad~\cite{smilkov2017smoothgrad}. Five commonly used neural network architectures including AlexNet~\cite{krizhevsky2012imagenet}, VGG-16~\cite{simonyan2014very}, ResNet-18~\cite{he2016deep}, DenseNet-169~\cite{huang2017densely}, and SqueezeNet~\cite{iandola2016squeezenet} and a state-of-the-art Swin Transformer model~\cite{liu2021swin} are used to showcase VOICE uncertainty. All models are pretrained and are publicly available from the PyTorch Torchvision library. The qualitative and quantitative results on clean data are presented in Sections~\ref{subsec:Qualitative_clean} and~\ref{subsec:Quantitative_clean} respectively while the results on challenging noisy data are presented in Sections~\ref{subsec:Qualitative_challenges} and~\ref{subsec:Quantitative_challenging} respectively.

\subsection{Qualitative results on clean data}
\label{subsec:Qualitative_clean}
We show qualitative results in Fig.~\ref{fig:Explanations_Correct} for a pretrained VGG-16 architecture~\cite{simonyan2014very}. GradCAM~\cite{selvaraju2017grad} and GradCAM++~\cite{chattopadhay2018grad} explanations are extracted from the last convolution layer to maximize the semantic content within the activations while preserving spatial locality. Guided Backpropagation~\cite{springenberg2014striving} and SmoothGrad~\cite{smilkov2017smoothgrad} are extracted by backpropagating the gradients to the input pixel space. Hence, GradCAM and GradCAM++ highlight semantic structures directly while Guided Backpropagation and SmoothGrad highlight pixels that collectively form the structure. In this paper, we choose these two classes of semantic and pixel explanations to illustrate our proposed uncertainty quantification.

Similar to Fig.~\ref{fig:Concept}, we differentiate between whether the network classified an image correctly or incorrectly. Note that all four considered explanations are \emph{post hoc} and hence, we have access to their classifications. We randomly choose three correctly classified images in the ImageNet~\cite{deng2009imagenet} validation set. These are a mosque, a bald eagle, and a motor scooter in Figs.~\ref{fig:Explanations_Correct}(a), (b), and (c) respectively. In Fig.~\ref{fig:Explanations_Correct}(a), the explanations provided by GradCAM and GradCAM++ both highlight the minaret. Additionally, GradCAM++ highlights the dome. The uncertainties of both GradCAM and GradCAM++ centre around their explanations with minor overlaps. This is more apparent in Fig.~\ref{fig:Explanations_Correct}(b), where the base of the beak of the bald eagle is a strong indicator of its class according to GradCAM. The uncertainty in GradCAM is noticeably centred around the base of the beak. This shows that GradCAM is certain in its explanation. However, this is not true for GradCAM++. A similar analysis holds for Fig.~\ref{fig:Explanations_Correct}(c) where the overlap between the explanation and its uncertainty is higher for GradCAM++ than GradCAM. In Section~\ref{subsec:Quantitative_clean}, we show that this observation can be generalized across multiple images. The results for Guided Backpropagation and SmoothGrad are less clear than their semantic counterparts due to the nature of pixelwise explanations. Guided backpropagation provides only a limited number of pixel highlights that make variance across that index across multiple contrastive explanations low. The uncertainties are highlighted in red. While the uncertainties are visible in SmoothGrad, it is more diffused compared to the uncertainties in GradCAM and GradCAM++. Also, more emphasis is placed on edges, which is a natural artifact of obtaining gradients from the input layer. Hence, by qualitatively observing visualizations of the uncertainties for Guided Backpropagation and SmoothGrad, no conclusions regarding the overlap between explanations and their uncertainties can be drawn. Quantitative evaluation in Section~\ref{subsec:Quantitative_clean} provides better results.

In Figs.~\ref{fig:Explanations_Correct}(d) and (e), we show explanations and their uncertainties when the network prediction is incorrect. The network predicts dogsled for Fig.~\ref{fig:Explanations_Correct}(d) and barbershop for Fig.~\ref{fig:Explanations_Correct}(e). In Fig.~\ref{fig:Explanations_Correct}(d), there is a noticeable large overlap between the explanations and uncertainties of GradCAM and GradCAM++. This indicates that the region that the network uses for classification is the same region it is uncertain about. This observation holds for SmoothGrad as well. Moreover, the uncertainties are dispersed in Figs.~\ref{fig:Explanations_Correct}(d) and (e), compared to the correctly classified images. 

\subsection{Quantitative results on clean data}
\label{subsec:Quantitative_clean}
In this section, we present a quantitative analysis to validate the overlap and dispersion hypotheses. We randomly select $500$ images from the ImageNet~\cite{deng2009imagenet} validation dataset and pass them through $6$ pretrained neural network architectures. The resulting predictions are analyzed based on being correct or incorrect in Table~\ref{tab:Results_All}. IoU and SNR metrics, differentiated between correct and incorrect predictions are shown. The third column in both IoU and SNR metrics titled $\%$ Difference in Table~\ref{tab:Results_All} is calculated as $\frac{(Wrong - Correct)}{Correct} \times 100$. Hence, it shows the $\%$ increase of IoU on wrong predictions from the correct predictions. This varies between different networks. For all networks except Swin Transformer, the IoU between each prediction's explanation and the uncertainty associated with that explanation is greater when the network is incorrect. Given that the network predicts incorrectly, it is more uncertain about the regions it is using to predict than if it were to have predicted correctly. This shows that the trust placed by the explanatory technique in its explanation is lesser since it is using the same features to explain its decision that it is uncertain about. Similarly, the SNR or dispersion metric is higher when the network predictions are incorrect across all networks and explanatory techniques.

\noindent\textbf{Effect of threshold $t$ on Swin transformer} In Table~\ref{tab:Results_Swin}, we analyze the effect of $t$ on the IoU metric on Swin Transformer. In Table~\ref{tab:Results_All}, $t$ is kept consistent at $0.1$ among all the networks. Increasing the value of $t$ increases the $\%$ difference in overlap between the correct and wrong predictions. This behavior is seen among both GradCAM and GradCAM++. Note that even at $0.1$, all CNN architectures in Table~\ref{tab:Results_All} show a positive $\%$ difference. On the other hand, Swin transformer has $-14.44\%$ difference indicating that the overlap on the correct predictions is higher than the incorrect predictions. This is because the explanations for GradCAM and GradCAM++ are extracted from a $7\times 7$ kernel which is interpolated to create a $224 \times 224$ image. This adds spurious overlaps between the explanations and uncertainties. Increasing threshold $t$ eliminates the spurious overlaps, thereby increasing the $\%$ difference as depicted in Table~\ref{tab:Results_Swin}.

\begin{figure*}[t!]
\begin{center}
\minipage{\textwidth}%
  \includegraphics[width=\linewidth]{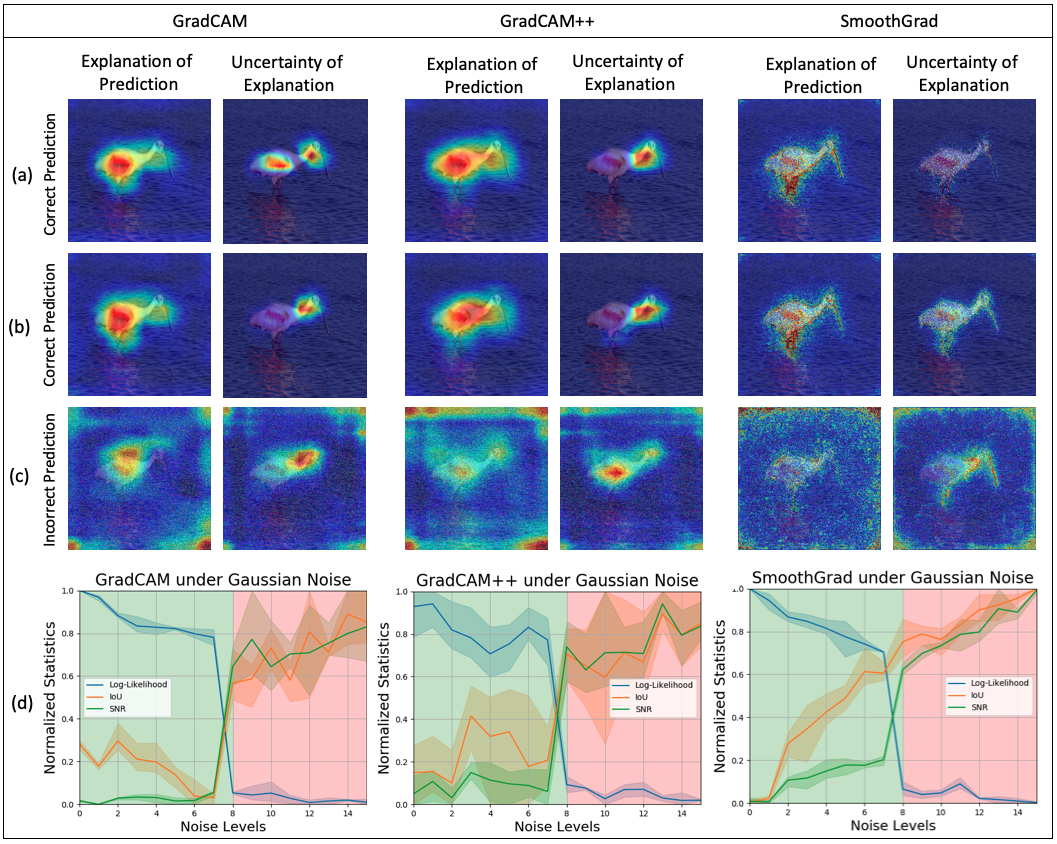}
\endminipage
\vspace{-1mm}
\caption{Explanations and uncertainty of a Spoonbill perturbed with AWGN at noise power levels of (a) 0, (b) 450, and (c) 11000, respectively. At 11000, the network incorrectly predicts a coral reef. (d) provides normalized log-likelihood, IoU and SNR metrics across $15$ noise levels. Shading indicates variance across three runs. Background green indicates correct prediction and red indicates incorrect prediction.}\label{fig:VGG_Qualiatative}
\end{center}
\end{figure*}

\input{Table/tab_Swin}

\subsection{Qualitative results on challenging data}
\label{subsec:Qualitative_challenges}
In this subsection, we analyze the qualitative and quantitative results on a single image of a spoonbill (ImageNet class $129$) across $15$ increasing levels of zero-mean Additive White Gaussian Noise (AWGN). Level 0 indicates the original image. Increasing levels indicate increase in noise power levels with Level 15 being the maximum considered noise. Note that after level 7, we increase the power level from $450$ to $7000$ so as to depict the trends between correct and incorrect predictions. We present these results for VGG-16~\cite{simonyan2014very} in Fig.~\ref{fig:VGG_Qualiatative}.

Figs.~\ref{fig:VGG_Qualiatative}(a), (b), and (c) show the explanations from GradCAM, GradCAM++, and SmoothGrad along with their VOICE uncertainties respectively. Row (a) in Fig.~\ref{fig:VGG_Qualiatative} are results on the original spoonbill image, row (b) are results on level 7 AWGN spoonbill, and row (c) are results on level 15 AWGN spoonbill. Row (d) in Fig.~\ref{fig:VGG_Qualiatative} shows the normalized quantitative metrics on IoU and SNR along with normalized log-likelihood on the \texttt{y-axis}, across the noise levels on the \texttt{x-axis}. Rows (a) and (b) spoonbills are correctly predicted as spoonbill by the VGG-16 network while the level 15 AWGN spoonbill in Row (c) is incorrectly predicted as coral reef. In row (d), the green background indicates that the network predicts the class correctly across levels 0 - 7, while levels 8 - 15 are incorrectly predicted and are represented by a red background. To obtain row (d) in Fig.~\ref{fig:VGG_Qualiatative}, we run 3 separate runs of the experiment at every noise level for each explanation of GradCAM, GradCAM++, and SmoothGrad. The epistemic uncertainty, represented by log-likelihood, follows the expected trajectory of gradual decrease with increase in noise levels. The steep fall between levels 7 and 8 is due to a large increase in noise power from $450$ to $7000$. Note that at lower levels of noise, the SNR and IoU of VOICE among all three explanations is lower. In SmoothGrad, the increase in IoU can be explained by looking at the VOICE explanations themselves. In row (b), the SmoothGrad VOICE is more prominent than row (a). Hence there is more overlap with the SmoothGrad explanation. In GradCAM++, the explanation and VOICE overlap remains consistent between rows (a) and (b). However, in GradCAM, the uncertainty \emph{decreases} with an increase in noise levels. This is reflected in a gradual decline in IoU in GradCAM row (d) when the background is green. Moreover, the variance across the $3$ runs is higher in IoU and SNR as compared to log-likelihood. This is because during each of the 3 experiments at every noise level, the added AWGN is random so that the predictions $P$ from the network are different across the $3$ runs. Hence, the VOICE uncertainty is asking \emph{'Why $P_r$, rather than Q?'} where $P_r$ is the prediction at run $r$ with $P_1 \neq P_2 \neq P_3$. Hence, the explanations and their corresponding uncertainties are different across the $3$ runs which is reflected in the higher variance. Note that in row (d) of Fig~\ref{fig:VGG_Qualiatative}, the values are normalized between $0$ and $1$. 
\begin{figure*}[t!]
\begin{center}
\minipage{\textwidth}%
  \includegraphics[width=\linewidth]{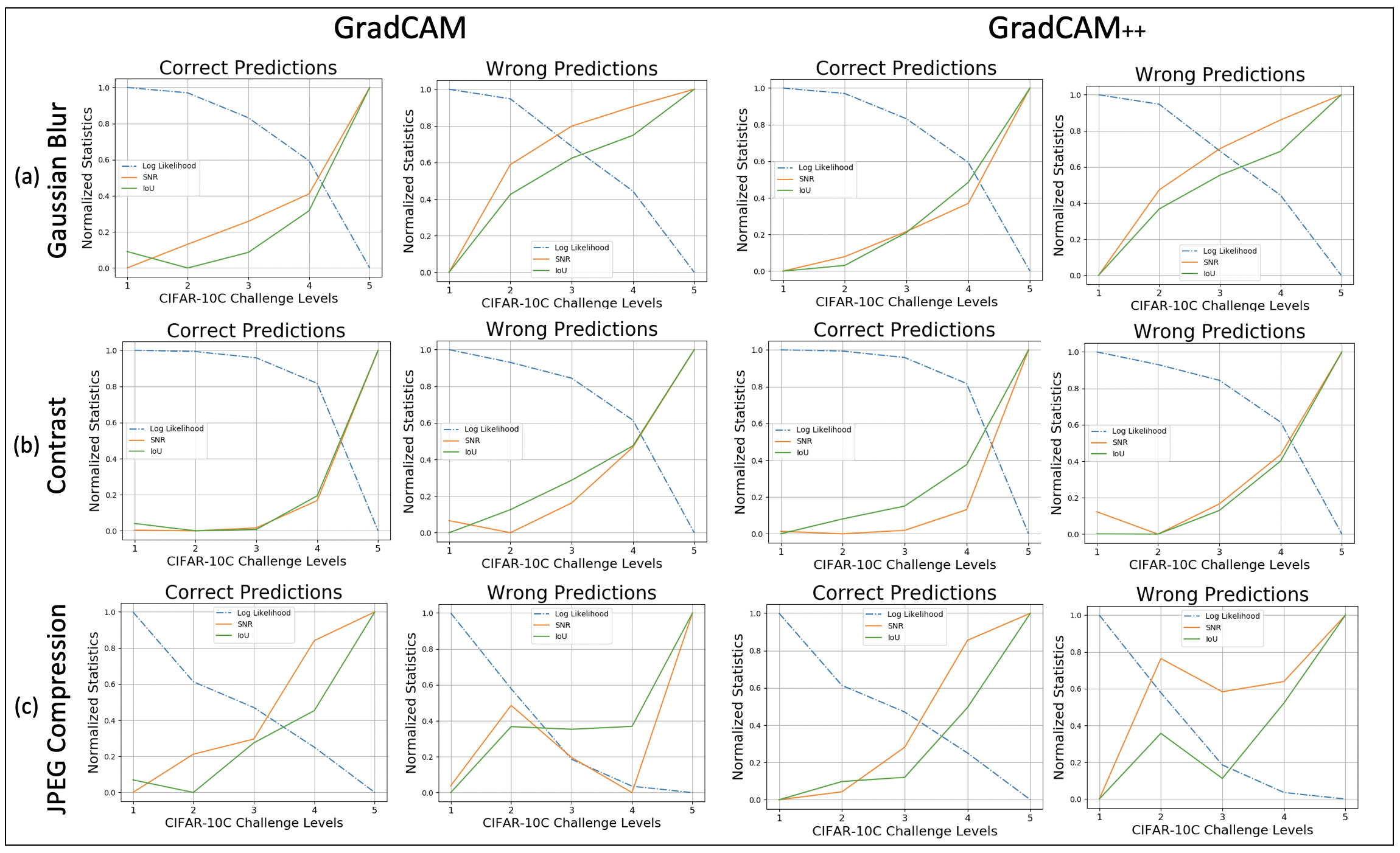}
\endminipage
\vspace{-1mm}
\caption{Quantitative results on three challenges from CIFAR-10C data on five progressively increasing challenge levels on the \texttt{x-axis}. First two columns are derived using GradCAM explanations and last two are derived from GradCAM++.}\label{fig:CIFAR-10C}
\end{center}
\end{figure*}

\subsection{Quantitative results on challenging data}
\label{subsec:Quantitative_challenging}
We use CIFAR-10C~\cite{hendrycks2019benchmarking} as the challenging dataset to benchmark the VOICE uncertainty metrics. CIFAR-10C provides $19$ environmental, acquisition, and noise challenges on CIFAR-10 data. The $10000$ images from CIFAR-10 testset are perturbed using $5$ progressively increasing levels of perturbations. Hence, each challenge has $50000$ images to benchmark VOICE uncertainty metrics. We obtain results on all $19$ challenges but show only three challenges in Fig.~\ref{fig:CIFAR-10C} due to space constraints. These three challenges including gaussian blur, contrast, and JPEG compression are chosen to showcase the diverse behaviors between IoU, SNR and log-likelihood. Fig.~\ref{fig:CIFAR-10C} consists of plots where \texttt{x-axis} depicts the challenge levels and \texttt{y-axis} plots normalized log-likelihood and VOICE metrics. The metrics are plotted for GradCAM and GradCAM++ explanations. Plots from each challenge and each explanation are divided based on correct or incorrect predictions. On contrast and gaussian blur challenges, the SNR and IoU VOICE metrics exhibit similar behavior as log-likelihood among both GradCAM and GradCAM++ explanations. However, in JPEG compression the behavior on challenge levels $3$ and $4$ between the three metrics deviates under incorrect predictions. This is observed in both GradCAM and GradCAM++ explanations. Among all $19$ challenges on CIFAR-10C, JPEG compression shows the largest deviation. Note that log-likelihood requires ground truth to quantify uncertainty while the proposed metrics do not.
\input{Table/tab_Compare}
\subsection{Comparing explanatory techniques}\label{subsec:Compare}
Table~\ref{tab:Results_All} provides IoU and SNR quantification of VOICE that validates the overlap and dispersion hypotheses. However, these results are insufficient to compare between explanations. For instance, Guided Backpropagation has the highest difference between correct and incorrect prediction IoU across all architectures, except Swin transformer. An incorrect conclusion from the preceding statement is that guided backpropagation is the best method (as compared against GradCAM, GradCAM++, and SmoothGrad) since it has less overlap between the explanation and its uncertainty. This is an incorrect conclusion since the values of IoU is not comparable between explanatory methodologies. On AlexNet, the IoU of Guided Backpropagation is $0.047$ while on the other explanations, IoU is around $0.5$. Visually, the low IoU value of Guided Backpropagation uncertainty manifests itself as mostly black regions in Fig.~\ref{fig:Explanations_Correct} with the uncertain regions highlighted within the red box to make them apparent.

Note that the variability in metric range is a common issue within uncertainty quantification literature. For instance, log-likelihood measure is unbounded and differs based on network, data, and predictions. Since there is no ground truth, uncertainty evaluation strategies utilize a set of samples rather than individual predictions for a pseudo-classification task. In~\cite{benkerttransitional}, the authors utilize their proposed uncertainty metric to classify between in-distribution and out-of-distribution data. Higher the classification accuracy, better the uncertainty metric. In~\cite{lee2023probing}, the authors use Area Under Accuracy Curve (AUAC) on CIFAR-10C dataset to showcase that their uncertainty metric is robust against increasing level of challenges. Ideally, AUAC must remain high, even when a network is presented with challenging data. We follow the same procedure as~\cite{lee2023probing} and provide the Area Under IoU and Area Under SNR Curves for each of the four explanations on three challenges used in Section~\ref{subsec:Quantitative_challenging}. From Fig.~\ref{fig:CIFAR-10C}, the IoU and SNR VOICE quantification must ideally be zero across increasing level of challenges. An explanation with its AUC closer to zero is a better explanation. We calculate the area under both IoU and SNR measures and present them in Table~\ref{tab:Results_Compare}. The best performing explanation for every challenge and metric is bolded. Note that on average, GradCAM has both IoU and SNR values closer to zero compared to the other explanations. Our proposed explanation overcomes the masking limitations in existing explanatory evaluation techniques to objectively quantify subjective explanations.

%% file: Table/tab_Results.tex
\begin{table*}[t]
\caption{Comparison among 4 explanatory techniques and 6 architectures on ImageNet dataset. Top-1 accuracy is shown. Ideally, the \% difference between IoU and SNR must be higher. The highlights in red are lower and explained in Table~\ref{tab:Results_Swin}.}
\label{tab:Results_All}
\begin{center}
\begin{footnotesize}
\begin{sc}
\begin{tabular}{lcccccccr}
\toprule
  \multirow{2}{*}{Architecture} & Accuracy & \multirow{2}{*}{Explanation} &  \multicolumn{3}{c}{IoU} & \multicolumn{3}{c}{SNR} \\

  & (\%) & & Correct & Wrong & \% Difference & Correct & Wrong & \% Difference  \\ 
  \midrule

\multirow{4}{*}{AlexNet~\cite{krizhevsky2012imagenet}} & \multirow{4}{*}{51.22} & GradCAM~\cite{selvaraju2017grad} & 0.5329 & 0.6242 & 17.13 & 1.0445 & 1.2952 & 24.00 \\

  & & GradCAM++~\cite{chattopadhay2018grad} & 0.5469 & 0.6732 & 23.09 & 1.0933 & 1.2620 & 15.44 \\

  & & Guided Backprop~\cite{springenberg2014striving} & 0.0469 & 0.0761 & 62.41 & 0.4686 & 0.4964 & 5.93 \\

  & & SmoothGrad~\cite{smilkov2017smoothgrad} & 0.4507 & 0.5904 & 30.99 & 1.0267 & 1.2010 & 16.97 \\

  \midrule

\multirow{4}{*}{SqueezeNet~\cite{iandola2016squeezenet}} & \multirow{4}{*}{54.17} & GradCAM~\cite{selvaraju2017grad} & 0.7163 & 0.7878 & 9.98 & 0.7869 & 0.9268 & 17.77 \\

  & & GradCAM++~\cite{chattopadhay2018grad} & 0.7535 & 0.8210 & 8.96 & 0.8455 & 0.9634 & 13.94 \\

  & & Guided Backprop~\cite{springenberg2014striving} & 0.1593 & 0.2166 & 35.98 & 0.3989 & 0.4280 & 7.29 \\
  
  & & SmoothGrad~\cite{smilkov2017smoothgrad} & 0.8556 & 0.9169 & 7.16 & 1.2719 & 1.4645 & 15.14 \\

  \midrule

\multirow{4}{*}{ResNet-18~\cite{he2016deep}} & \multirow{4}{*}{65.23} & GradCAM~\cite{selvaraju2017grad} & 0.7232 & 0.8300 & 14.77 & 1.4019 & 1.7366 & 26.51 \\

  & & GradCAM++~\cite{chattopadhay2018grad} & 0.7110 & 0.8189 & 15.18 & 1.2482 & 1.4113 & 13.07 \\

  & & Guided Backprop~\cite{springenberg2014striving} & 0.0322 & 0.0446 & 38.28 & 0.4056 & 0.4309 & 6.26 \\
  
  & & SmoothGrad~\cite{smilkov2017smoothgrad} & 0.3626 & 0.4620 & 27.43 & 0.9306 & 1.0755 & 15.57 \\

  \midrule

  \multirow{4}{*}{VGG-16~\cite{simonyan2014very}} & \multirow{4}{*}{63.57} & GradCAM~\cite{selvaraju2017grad} & 0.3939 & 0.5234 & 32.89 & 0.7877 & 0.9536 & 21.06 \\

  & & GradCAM++~\cite{chattopadhay2018grad} & 0.3905 & 0.5474 & 40.17 & 0.7923 & 0.9653 & 21.83 \\

  & & Guided Backprop~\cite{springenberg2014striving} & 0.0328 & 0.0493 & 50.31 & 0.3683 & 0.4004 & 8.73 \\

  & & SmoothGrad~\cite{smilkov2017smoothgrad} & 0.3083 & 0.4493 & 45.73 & 0.8736 & 1.0961 & 25.47 \\

  \midrule

  \multirow{4}{*}{DenseNet-169} & \multirow{4}{*}{67.83} & GradCAM~\cite{selvaraju2017grad} & 0.5892 & 0.7860 & 33.39 & 1.1521 & 1.4572 & 26.48 \\

  & & GradCAM++~\cite{chattopadhay2018grad} & 0.5950 & 0.7645 & 28.48 & 1.1551 & 1.3261 & 14.8 \\

  & & Guided Backprop~\cite{springenberg2014striving} & 0.031 & 0.0522 & 68.54 & 0.3959 & 0.4305 & 8.74 \\

  & & SmoothGrad~\cite{smilkov2017smoothgrad} & 0.2735 & 0.3962 & 44.87 & 0.7937 & 0.9870 & 24.35 \\

  \midrule

  \multirow{4}{*}{Swin Base~\cite{liu2021swin}} & \multirow{4}{*}{80.61} & GradCAM~\cite{selvaraju2017grad} & 0.7151 & 0.7524 & 5.22 & 0.8655 & 1.0123 & 16.96 \\

& & GradCAM++~\cite{chattopadhay2018grad} & 0.2111 & 0.1806 & \textcolor{red}{-14.44} & 0.9608 & 1.0809 & 12.50 \\

  & & Guided Backprop~\cite{springenberg2014striving} & 0.0606 & 0.0524 & \textcolor{red}{-13.49} & 0.2556 & 0.2708 & 5.93 \\
  
  & & SmoothGrad~\cite{smilkov2017smoothgrad} & 0.3394 & 0.3723 & 9.69 & 1.1868 & 1.2606 & 6.22 \\
  
  \bottomrule
  
\end{tabular}
\end{sc}
\end{footnotesize}
\end{center}
\end{table*}

%% file: Table/tab_Swin.tex
\begin{table}[t]
\caption{Results of IoU on Swin Transformer with a change in threshold $t$.}
\label{tab:Results_Swin}
\begin{center}
\begin{tiny}
\begin{sc}
\begin{tabular}{lcccccr}
\toprule
Threshold &  \multicolumn{3}{c}{GradCAM} & \multicolumn{3}{c}{GradCAM++} \\

  $t$ & Correct & Wrong & \% Difference & Correct & Wrong & \% Difference  \\ 
  \midrule
  
  0.1 & 0.7151 & 0.7524 & 5.22 & 0.2111 & 0.1806 & \textcolor{red}{-14.44} \\

  0.3 & 0.3604 & 0.3756 & 4.21 & 0.1472 & 0.1361 & \textcolor{red}{-7.55} \\

  0.4 & 0.1970 & 0.2103 & 6.77 & 0.0994 & 0.1033 & 3.90 \\

  0.5 & 0.0930 & 0.1055 & 13.44 & 0.0592 & 0.0737 & 24.52 \\

  0.6 & 0.0384 & 0.0485 & 26.29 & 0.0347 & 0.0531 & 53.15 \\

  0.7 & 0.0830 & 0.1206 & 45.34 & 0.0210 & 0.0364 & 73.04 \\
  
  \bottomrule
  
\end{tabular}
\end{sc}
\end{tiny}
\end{center}
\end{table}

%% file: Table/tab_Compare.tex
\begin{table}[t]
\caption{Comparing explanatory techniques using AUC scores on CIFAR-10C dataset. The best performing explanation for every challenge and metric is bolded.}
\label{tab:Results_Compare}
\begin{center}
\begin{sc}
\begin{tabular}{lccr}
\toprule
Challenges &  Explanation & IoU & SNR \\
  \midrule
  
  \multirow{4}{*}{Gaussian Blur} & GradCAM & \textbf{0.24} & 0.31 \\
  & GradCAM++ & 0.27 & 0.29\\
  & Guided Backprop & 0.46 & \textbf{0.19}\\
  & SmoothGrad & 0.32 & 0.49 \\

  \midrule 

  \multirow{4}{*}{Contrast} & GradCAM & \textbf{0.17} & 0.17\\
  & GradCAM++ & 0.19 & \textbf{0.16}\\
  & Guided Backprop & 0.22 & 0.29 \\
  & SmoothGrad & 0.42 & 0.23 \\

  \midrule 

  \multirow{4}{*}{JPEG Compression} & GradCAM & 0.30 & \textbf{0.33} \\
  & GradCAM++ & 0.31 & 0.39\\
  & Guided Backprop & \textbf{0.27} & 0.41 \\
  & SmoothGrad & \textbf{0.27} & 0.47 \\
  
  \bottomrule
  
\end{tabular}
\end{sc}
\end{center}
\end{table}

%% file: Sections/6_Discussion.tex
\section{Discussion}
\label{sec:Discussion}

\begin{figure}[ht!]
\begin{center}
\minipage{\textwidth}%
  \includegraphics[width=0.49\linewidth]{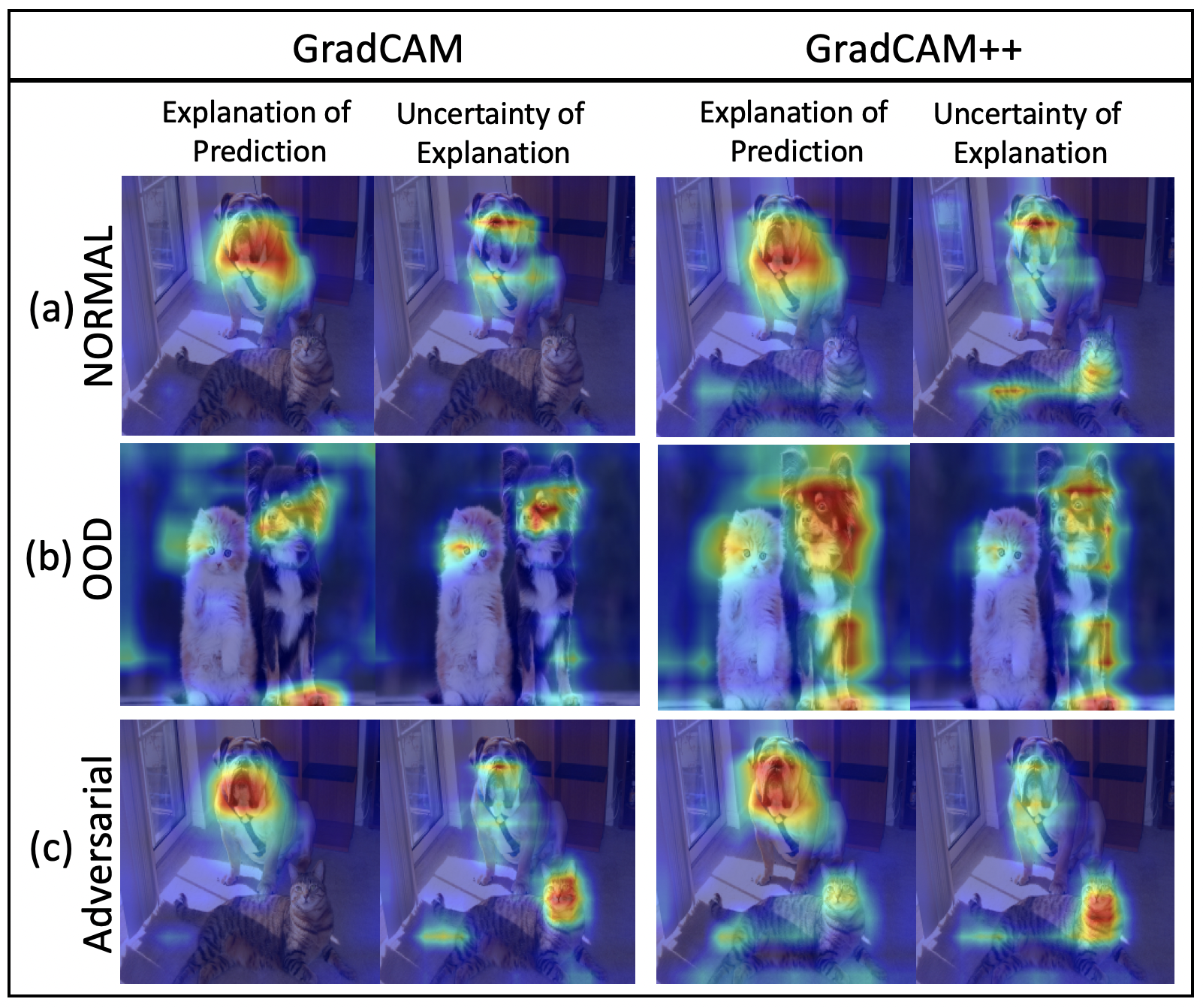}
\endminipage
\vspace{-1mm}
\caption{Visualization of explanations and their corresponding uncertainties under (a) normal, (b) out-of-distribution, and (c) adversarial data.}\label{fig:OOD}
\end{center}
\end{figure}

\subsection{Takeaways}
\paragraph{Non-gradient based explanation uncertainty}
The current work of obtaining uncertainty of explanations is applicable to only gradient-based attribution methods. The proposed VOICE methodology is a white-box method, i.e. it requires access to all the network parameters and its gradients to induce uncertainty. While gradient-based methods are a popular means of visualizing feature attributions, there are other methods that provide explanations. These include TCAV~\cite{kim2018interpretability}, LIME~\cite{ribeiro2016should}, and Graph-CNN~\cite{zhang2018interpreting} among others. Future work on obtaining method-based uncertainty must focus on non-gradient explanatory techniques as well. 

\paragraph{Objective measures for subjective explanations}
The theoretical analysis of predictive uncertainty shows that the current methodology of objectively evaluating explanations only, partly, reduces uncertainty. Masking images and decreasing the $V[E(y|S_x)]$ term ignores the second term $E[V(y|S_x)]$ in Eq.~\ref{eq:Decomposition}. A simple way of decreasing $V[E(y|S_x)]$ is by having bigger masks. But this invariably leads to a larger area of uncertainty and hence larger $E[V(y|S_x)]$. Any objective evaluation methodology that relies on masking suffers from the same issue. Hence, either uncertainty must also be reported along with accuracy evaluations or the proposed area under SNR and IoU curves must be reported.

\paragraph{VOICE uncertainty on network-targeted challenges} 
In this paper, we used challenging data from CIFAR-10C dataset where perturbations are obtained based on acquisition challenges like contrast, shot noise, motion blur, environmental challenges like snow, rain, fog, as well as noise characteristics. These are natural challenges that have shown to affect the recognition performance of neural networks. However, neural network specific challenges like adversarial images and out-of-distribution images that are not present in the training domain also affect the network performance. We show two such examples in Fig.~\ref{fig:OOD}. VGG-16 architecture is used to obtain all explanations and uncertainties. The out-of-distribution (OOD) image in Fig.~\ref{fig:OOD}(b) is correctly classified despite not being a part of ImageNet. There is a larger overlap between the explanations and uncertainties among both explanations in Fig.~\ref{fig:OOD}(b) compared to Fig.~\ref{fig:OOD}(a). Note that the prediction is correct for the OOD image unlike in Section~\ref{subsec:Quantitative_clean} where a larger overlap is observed when the prediction is incorrect. The adversarial image in Fig.~\ref{fig:OOD}(c) is derived using I-FGSM method~\cite{goodfellow2014explaining}. In Fig.~\ref{fig:OOD}(a), the image is recognized correctly as a bull-mastiff. In Fig.~\ref{fig:OOD}(c), the image is recognized incorrectly as a boxer. However, both GradCAM and GradCAM++ highlight the features of the dog in Fig.~\ref{fig:OOD}(a) and Fig.~\ref{fig:OOD}(c) to explain different predictions. In contrast, the uncertainties of GradCAM and GradCAM++ in Fig.~\ref{fig:OOD}(c) highlight the face of the cat while the uncertainties of GradCAM and GradCAM++ in Fig.~\ref{fig:OOD}(a) highlight the features around the explanation. Hence, there is a spatial difference in occurrence of uncertainties while there is no change in the explanations themselves. This indicates that while explanations by themselves may not be sufficient to detect adversarial images, VOICE uncertainty may be capable. However, the overlap metric is insufficient here and future work must focus on new VOICE measures on adversarial and OOD image detection.

\subsection{Conclusion}
In this paper, we show that all visual explanatory techniques have some inherent uncertainty associated with them. We provide a theoretical analysis based on predictive uncertainty and observe that existing evaluation techniques for explanations only partially reduce predictive uncertainty. We propose a methodology based on contrastive explanations to visualize the remaining predictive uncertainty. In doing so, we observe that under incorrect predictions, neural networks are often uncertain about the same features that they use to make predictions. Moreover, the dispersion of uncertainty as well as the overlap of an explanation and its uncertainty can be used to quantify the explanation's uncertainty. These quantities exhibit similar behavior as log-likelihood, a measure of epistemic uncertainty.